\documentclass[10pt]{article}

\usepackage[T1]{fontenc}
\usepackage[utf8]{inputenc}
\usepackage{lmodern}
\usepackage[margin=1in]{geometry}
\usepackage{amsmath,amssymb}
\usepackage{graphicx}
\usepackage{booktabs}
\usepackage{multirow}
\usepackage{arydshln}
\usepackage{adjustbox}
\usepackage{xcolor}
\usepackage{microtype}
\usepackage[section]{placeins}
\usepackage[hidelinks]{hyperref}

\newcommand{\best}[1]{\textcolor{red}{\textbf{#1}}}
\newcommand{\second}[1]{\textcolor{blue}{\textbf{#1}}}

\title{Analyzing Multi-Layer Visual Representations of\\Large Vision-Language Models for Anomaly Detection}

\author{
  Satoshi Hashimoto\quad Hitoshi Nishimura\quad Mori Kurokawa \\[4pt]
  KDDI Research, Inc. \\
  \texttt{st-hashimoto@kddi.com}
}

\date{}

\begin{document}

\maketitle

\begin{abstract}
In this paper, we analyze multi-layer visual representations in large vision-language models (LVLMs) and examine the effectiveness of layer-wise features and multi-layer fusion for anomaly detection.
Representative LVLMs connect visual encoder features to a language model through a projector.
In this design, deep visual encoder features carry high-level semantics and alignment with text representations, whereas shallow and intermediate layers retain information relevant to vision-centric tasks, such as local appearance and fine-grained shape.
It is therefore important to choose layer representations and fusion methods according to the nature of the downstream task.
Focusing on anomaly detection, we compare layer-wise features and multi-layer fusion across multiple supervision levels and a variety of classifiers, and examine from several perspectives how multi-layer visual representations affect downstream performance.
By analyzing the resulting performance differences, we further discuss the contribution of each layer representation to anomaly detection and provide empirical insights into representation selection for vision-centric tasks.
\end{abstract}

\section{Introduction}

Large vision-language models (LVLMs) have recently shown strong performance on a wide range of vision-language tasks, including image and video understanding.
Representative LVLMs map visual features from a pretrained visual encoder into the embedding space of a large language model (LLM) through a projector, and process the resulting visual tokens as part of the language model's input sequence; this design has been adopted on empirical grounds~\cite{liu2023visual_instruction_tuning,lin2025multi_layer_visual_feature_fusion}.
Final-layer features are a natural choice because they carry high-level semantics and are aligned with text representations.
However, shallow and intermediate layers of the visual encoder may retain information useful for vision-centric tasks, such as local appearance, fine-grained shape, and object layout.
From this perspective, recent studies have revisited visual layer selection and multi-layer visual feature fusion in LVLMs~\cite{chen2025look_shallower,lin2025multi_layer_visual_feature_fusion}.
Chen et al.\ reported that shallow and intermediate features can outperform deep features on vision-centric tasks such as counting, position estimation, and object localization~\cite{chen2025look_shallower}.
Lin et al.\ systematically compared multi-layer visual feature fusion strategies in LVLMs and showed that both layer selection and the fusion method affect downstream performance~\cite{lin2025multi_layer_visual_feature_fusion}.

\begin{figure}[t]
\centering
\includegraphics[width=0.6\linewidth]{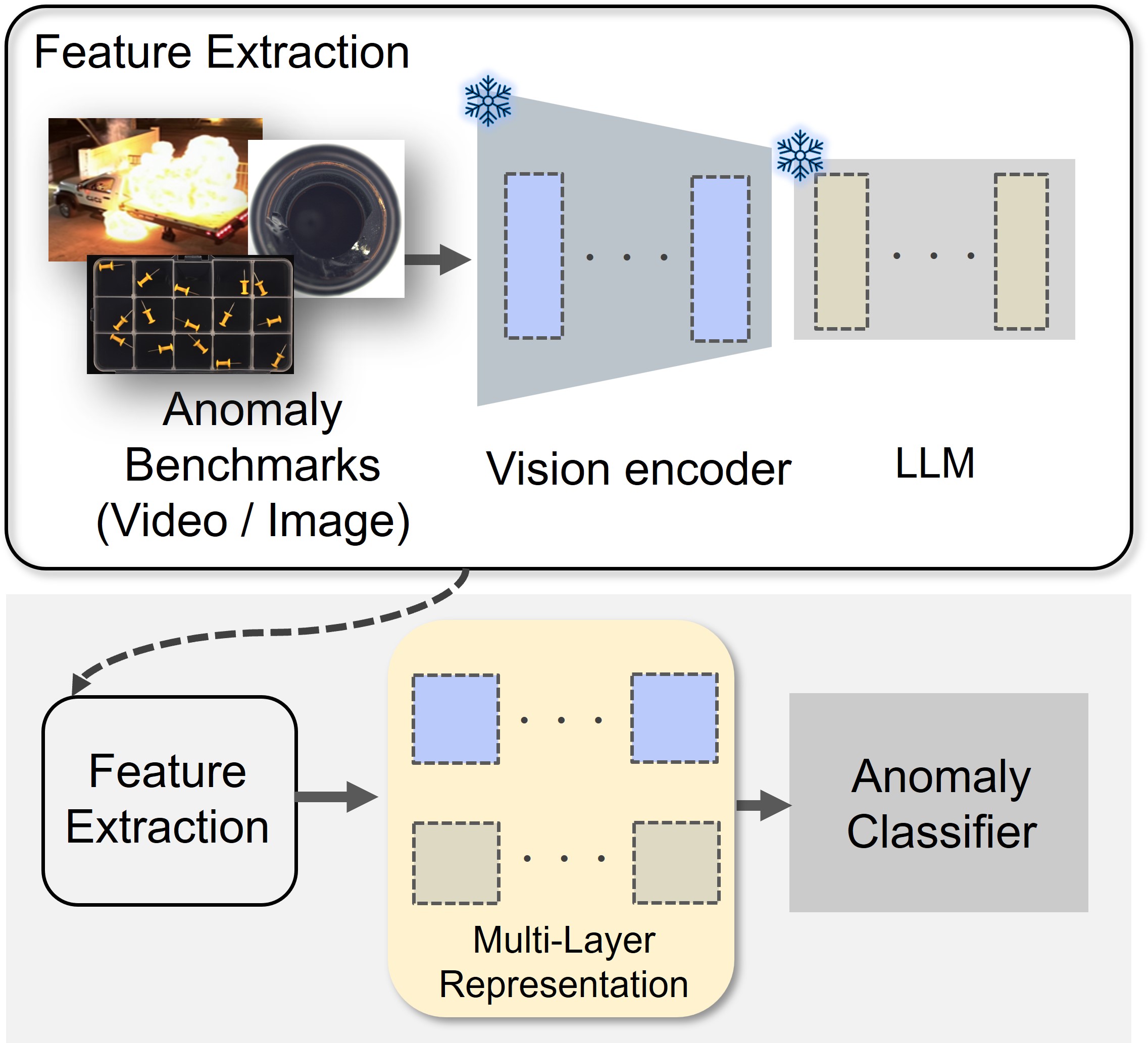}
\caption{Overview of the multi-layer visual representation analysis.}
\label{fig:overview}
\end{figure}

However, existing studies mainly examine a few representative layers of the visual encoder, and comparisons across a broad range of internal layers, including both the visual encoder and the language model, remain limited.
Moreover, evaluations have centered on typical vision-language benchmarks such as VQA, and insights into layer selection for anomaly detection, which is important in practical applications, have not been sufficiently organized.
It is also unknown how differences in supervision granularity (supervised, weakly supervised, and one-class settings) and the properties of the classifier affect which layer representations are effective.

In this work, we analyze the effectiveness of layer-wise features and multi-layer fusion for anomaly detection, targeting the multi-layer representations of LVLMs (Fig.~\ref{fig:overview}).
Specifically, we extract features from each layer of the visual encoder and the language model and compare their performance on image and video anomaly detection.
We further evaluate supervised, weakly supervised, and one-class settings on the same datasets and observe how the effective layer representations change with the supervision signal.
This allows us to revisit the conventional reliance on final-layer features and to offer practical guidelines for using the internal representations of LVLMs in vision-centric tasks.
Our experiments show that, in contrast to prior claims that shallow/intermediate layers and multi-layer fusion are effective~\cite{chen2025look_shallower,lin2025multi_layer_visual_feature_fusion}, the effective representation for anomaly detection depends on the properties of the normal distribution and on the supervision, and can be a single layer or a deep LLM-side feature.

\section{Related Work}

\paragraph{Multi-layer representation analysis in LVLMs.}
Lin et al.\ analyzed layer selection and fusion strategies for visual encoder features in LVLMs, and reported that selecting representative features from shallow, intermediate, and deep layers is beneficial, that adding shallow features helps on tasks requiring fine details, and that direct fusion at the input stage yields stable performance~\cite{lin2025multi_layer_visual_feature_fusion}.
Chen et al.\ evaluated the relationship between visual encoder depth and task performance on multiple benchmarks, and showed that deep features are effective for tasks requiring language alignment such as OCR, whereas shallow and intermediate features are effective for vision-centric tasks such as counting, position estimation, and localization~\cite{chen2025look_shallower}.
Mitra et al.\ showed that internal attention-head representations of generative LVLMs enable few-shot vision-language classification without relying on output tokens~\cite{mitra2025few_shot_lmm_features}.
These studies demonstrate the usefulness of multi-layer representations in LVLMs, but they mainly evaluate general vision-language benchmarks and do not examine layer-wise representations or supervision granularity in image and video anomaly detection.

\paragraph{Anomaly detection and LVLMs.}
In image anomaly detection, methods based on deviations from the distribution of normal features are widely used; PatchCore, which relies on a memory bank of normal patch features, is a representative example~\cite{roth2022patchcore}.
LogSAD handles structural and logical anomalies in a training-free manner using vision and language foundation models~\cite{zhang2025logsad}.
Video anomaly detection must handle deviations in actions and temporal context in addition to appearance anomalies, and is formulated according to the granularity of the supervision signal, e.g., fully supervised, weakly supervised, or trained on normal data only.
SteerVAD improves video anomaly detection by selecting anomaly-relevant attention heads inside the LLM of a frozen LVLM and rectifying the latent representation manifold~\cite{cai2026steervad}.
This method is related to ours in that it uses the internal representations of an LVLM for video anomaly detection, but it mainly targets LLM-side attention heads and does not address layer-wise features of the visual encoder or multi-layer fusion.
In this work, we analyze the relationship between layer selection and the supervision signal in anomaly detection, covering multi-layer representations that include the visual encoder.

\section{Analysis Method}

We analyze how the internal representations of an LVLM affect anomaly detection from the viewpoints of layer depth, supervision granularity, classifier type, and target benchmark.
Existing studies compare shallow, intermediate, and deep layers of the visual encoder as representative layers~\cite{chen2025look_shallower,lin2025multi_layer_visual_feature_fusion}, and we follow this approach for the visual encoder.
For a broader comparison, we also include the projected features after the projector and the internal features of the LLM.

\paragraph{Feature extraction and classifiers.}
It has been reported that LVLMs can be used not only for generating responses but also as sources of discriminative features from their internal representations~\cite{mitra2025few_shot_lmm_features}.
Accordingly, we feed the features extracted from each layer of a frozen LVLM into a downstream classifier and obtain detection results, comparing how each layer representation affects anomaly detection.
Let $\mathbf{x}$ be an input sample and let the token feature matrix of the $l$-th layer be
\begin{equation}
\mathbf{H}_{l}(\mathbf{x})
=
[\mathbf{h}_{l}^{(1)}(\mathbf{x}),\ldots,\mathbf{h}_{l}^{(N_l)}(\mathbf{x})]^{\top}
\in \mathbb{R}^{N_l \times D_l},
\end{equation}
where $N_l$ is the number of tokens, $D_l$ is the feature dimension, and $\mathbf{h}_{l}^{(n)}(\mathbf{x})\in\mathbb{R}^{D_l}$ is the feature vector of the $n$-th token.
Each layer feature is aggregated by averaging over tokens:
\begin{equation}
\mathbf{z}_{l}(\mathbf{x})
=
\frac{1}{N_l}
\sum_{n=1}^{N_l}
\mathbf{h}_{l}^{(n)}(\mathbf{x})
\in \mathbb{R}^{D_l}.
\end{equation}
When a single layer is used, we set
\begin{equation}
\boldsymbol{\phi}_{l}(\mathbf{x})
=
\mathbf{z}_{l}(\mathbf{x}).
\end{equation}
When a set of layers $\mathcal{L}$ is used, the averaged feature vectors of the layers are concatenated:
\begin{equation}
\boldsymbol{\phi}_{\mathcal{L}}(\mathbf{x})
=
\mathrm{Concat}
\left(
\mathbf{z}_{l}(\mathbf{x})
\mid
l\in\mathcal{L}
\right).
\end{equation}
The extracted feature $\boldsymbol{\phi}_{\mathcal{L}}(\mathbf{x})$ is fed into a classifier $f_{\theta}$ to obtain the anomaly score
\begin{equation}
s(\mathbf{x})
=
f_{\theta}
\left(
\boldsymbol{\phi}_{\mathcal{L}}(\mathbf{x})
\right),
\end{equation}
where $s(\mathbf{x})\in\mathbb{R}$ is the degree of anomaly of the input sample $\mathbf{x}$.

\paragraph{Benchmarks and supervision settings.}
We compare the effectiveness of layer representations on several major image and video anomaly detection benchmarks.
We consider three supervision settings: fully supervised, weakly supervised, and one-class.
In the fully and weakly supervised settings, a classifier learns the anomaly score; in the one-class setting, the deviation from normal features is used as the anomaly score.

\section{Experiments}

\subsection{Setup}

\paragraph{Multi-layer representation settings.}
We compare anomaly detection performance using layer-wise features extracted from a frozen LVLM, Qwen2.5-VL-7B~\cite{qwen2025qwen25vl}.
We denote the feature of the $k$-th visual encoder layer by V-$k$, the feature after the projector by V-Proj, and the image-token feature at the $k$-th LLM layer by L-$k$.
Following the analysis of Chen et al.~\cite{chen2025look_shallower}, we evaluate as single-layer features the visual layers 7, 15, 23, and 31 and the projector output, and the LLM layers 6, 13, 20, and 27.
As multi-layer features, we use, on the visual side, a shallow group V-Shallow (layers 7 and 15), a deep group V-Deep (layers 23 and 31 and the projector output), and an all-layer group V-All; on the LLM side, a shallow group L-Shallow (layers 6 and 13), a deep group L-Deep (layers 20 and 27), and an all-layer group L-All.

\paragraph{Benchmarks and classifiers.}
For video anomaly detection, we use UBnormal~\cite{acsintoae2022ubnormal} and UCSD Ped2~\cite{mahadevan2010anomaly} and evaluate frame-level AUROC.
For image anomaly detection, we use MVTec AD~\cite{bergmann2019mvtec_ad} and MVTec LOCO AD~\cite{bergmann2022mvtec_loco_ad} and evaluate image-level AUROC.
The classifier consists of a Transformer encoder and a linear output layer.
In the supervised setting, the classifier learns anomaly scores from the input features using image-level labels.
In the weakly supervised setting, scores are computed from sequences of segment features and the classifier is trained with video-level labels based on multiple-instance learning.
In the one-class setting, we compare a Transformer reconstructor (Recon), a Transformer predictor (Pred), and PatchCore~\cite{roth2022patchcore}.
The anomaly score is the reconstruction error of the input for the reconstructor, the prediction error for the predictor, and the nearest-neighbor distance to the normal memory bank for PatchCore; for PatchCore, token averaging is not applied in order to retain patch information.

\subsection{Overall Results}

Table~\ref{tab:summary_best_layer} lists the best feature and its AUROC for each benchmark.
Overall, multi-layer visual features were strong, but their advantage was not consistent, and single-layer or LLM-side features were best in some cases.
This suggests that, for anomaly detection, the multi-layer fusion strategy for visual encoders proposed by Lin et al.~\cite{lin2025multi_layer_visual_feature_fusion} is not optimal.
In particular, in the one-class setting, a single layer of either the LLM or the visual encoder was sometimes best.
We attribute this to the fact that multi-layer fusion can complicate the normal distribution, whereas single-layer features represent the normal manifold more simply and stably, which suits anomaly scoring based on distance, reconstruction, or prediction.
Therefore, when using LVLM features for anomaly detection, it is important not to presuppose multi-layer fusion, but to select layers, including LLM-side features, according to the supervision setting and the classifier.


\begin{table}[!htbp]
\centering
\caption{Best feature and its AUROC on video and image anomaly detection benchmarks.}
\label{tab:summary_best_layer}
\small
\begin{tabular}{lllccc}
\toprule
Task & Benchmark & Protocol & Method & Best feature & Best AUROC \\
\midrule
\multirow{8}{*}{Video AD}
& \multirow{4}{*}{UBnormal}
& Supervised        & Cls.   & V-All  & \textbf{70.92} \\
& & Weakly-supervised & Cls.   & V-Deep & \textbf{68.55} \\
& & One-class         & Pred.  & L-27   & \textbf{57.68} \\
& & One-class         & Recon. & L-20   & \textbf{60.10} \\
\cmidrule(lr){2-6}
& \multirow{4}{*}{UCSD Ped2}
& Supervised        & Cls.   & L-13  & \textbf{78.89} \\
& & Weakly-supervised & Cls.   & V-All & \textbf{76.82} \\
& & One-class         & Pred.  & V-15  & \textbf{71.51} \\
& & One-class         & Recon. & V-All & \textbf{96.45} \\
\midrule
\multirow{4}{*}{Image AD}
& \multirow{2}{*}{MVTec AD}
& One-class         & Recon.    & V-Proj & \textbf{85.60} \\
& & One-class         & PatchCore & V-15  & \textbf{94.01} \\
\cmidrule(lr){2-6}
& \multirow{2}{*}{MVTec LOCO AD}
& One-class         & Recon.    & V-All & \textbf{79.92} \\
& & One-class         & PatchCore & V-23  & \textbf{71.40} \\
\bottomrule
\end{tabular}
\end{table}

\begin{table}[!htbp]
\centering
\caption{Layer-wise AUROC of the visual encoder and the LLM on UBnormal.
Red and blue indicate the best and second-best results, respectively.
V-$k$ denotes the $k$-th visual encoder layer, V-Proj the feature after the projector, and L-$k$ the image-token feature at the $k$-th LLM layer.
Shallow, Deep, and All denote the multi-layer features of the shallow, deep, and all-layer groups, respectively.}
\label{tab:ubnormal_layer_sweep}
\resizebox{\textwidth}{!}{%
\begin{tabular}{lccccccccccccccc}
\toprule
 & \multicolumn{8}{c}{Vision} & \multicolumn{7}{c}{LLM} \\
\cmidrule(lr){2-9}\cmidrule(lr){10-16}
Method & V-7 & V-15 & V-23 & V-31 & V-Proj & V-Shallow & V-Deep & V-All & L-06 & L-13 & L-20 & L-27 & L-Shallow & L-Deep & L-All \\
\midrule
Supervised & 60.24 & 63.38 & 66.53 & 66.80 & \second{70.23} & 61.63 & 68.73 & \best{70.92} & 62.20 & 60.49 & 66.65 & 65.45 & 63.11 & 64.53 & 65.36 \\
\hdashline
Weakly-supervised & 54.90 & 53.32 & 58.20 & 60.24 & \second{67.48} & 54.96 & \best{68.55} & 62.46 & 53.95 & 53.92 & 60.16 & 61.55 & 50.77 & 60.25 & 59.42 \\
\hdashline
One-Class-Pred & 53.99 & 55.37 & 55.44 & 55.47 & 54.08 & 55.37 & 54.39 & 54.63 & 55.12 & 54.92 & 57.21 & \best{57.68} & 55.14 & \second{57.54} & 56.28 \\
\hdashline
One-Class-Recon & 56.46 & 58.00 & 58.57 & 57.71 & 55.98 & 57.25 & 56.72 & 56.90 & 57.90 & 56.87 & \best{60.10} & 59.53 & 57.12 & \second{59.99} & 58.78 \\
\bottomrule
\end{tabular}
}
\end{table}

\begin{table}[!htbp]
\centering
\caption{Layer-wise AUROC on UCSD Ped2.}
\label{tab:ucsdped2_layer_sweep}
\resizebox{\textwidth}{!}{%
\begin{tabular}{lccccccccccccccc}
\toprule
 & \multicolumn{8}{c}{Vision} & \multicolumn{7}{c}{LLM} \\
\cmidrule(lr){2-9}\cmidrule(lr){10-16}
Method & V-7 & V-15 & V-23 & V-31 & V-Proj & V-Shallow & V-Deep & V-All & L-06 & L-13 & L-20 & L-27 & L-Shallow & L-Deep & L-All \\
\midrule
Supervised & 72.95 & 69.98 & \second{78.30} & 72.50 & 75.58 & 73.44 & 77.33 & 74.10 & 72.29 & \best{78.89} & 73.23 & 74.53 & 76.44 & 76.49 & 77.58 \\
\hdashline
Weakly-supervised & 74.25 & 69.15 & 75.08 & 67.29 & 72.93 & 75.26 & \second{76.62} & \best{76.82} & 67.54 & 74.04 & 71.98 & 70.17 & 69.70 & 72.59 & 70.01 \\
\hdashline
One-Class-Pred & 58.46 & \best{71.51} & \second{71.23} & 58.83 & 51.05 & 64.63 & 58.32 & 59.24 & 59.74 & 68.02 & 62.34 & 57.50 & 69.63 & 60.28 & 62.33 \\
\hdashline
One-Class-Recon & 90.17 & 92.28 & 93.77 & \second{96.06} & 93.16 & 95.44 & 95.60 & \best{96.45} & 91.38 & 95.51 & 94.42 & 90.85 & 94.47 & 93.99 & 95.89 \\
\bottomrule
\end{tabular}
}
\end{table}

\begin{table}[!htbp]
\centering
\caption{Layer-wise AUROC on MVTec AD.}
\label{tab:mvtecad_layer_sweep_summary}
\resizebox{\textwidth}{!}{%
\begin{tabular}{llccccccccccccccc}
\toprule
\multicolumn{2}{c}{} & \multicolumn{8}{c}{Vision} & \multicolumn{7}{c}{LLM} \\
\cmidrule(lr){3-10}\cmidrule(lr){11-17}
Method & Metric & V-7 & V-15 & V-23 & V-31 & V-Proj & V-Shallow & V-Deep & V-All & L-06 & L-13 & L-20 & L-27 & L-Shallow & L-Deep & L-All \\
\midrule
One-Class-Recon & Overall & 63.75 & 70.81 & 79.81 & \second{84.80} & \best{85.60} & 68.46 & 84.23 & 81.27 & 80.90 & 80.67 & 82.97 & 81.40 & 81.42 & 82.95 & 82.29 \\
\midrule
One-Class PatchCore & Overall & 80.96 & \best{94.01} & \second{93.53} & 82.32 & 79.43 & 91.83 & 82.46 & 87.57 & 75.60 & 81.39 & 80.24 & 79.14 & 78.50 & 78.54 & 80.14 \\
\bottomrule
\end{tabular}
}
\end{table}

\begin{table}[!htbp]
\centering
\caption{Layer-wise AUROC on MVTec LOCO AD.}
\label{tab:mvtecloco_layer_sweep_summary}
\resizebox{\textwidth}{!}{%
\begin{tabular}{llccccccccccccccc}
\toprule
\multicolumn{2}{c}{} & \multicolumn{8}{c}{Vision} & \multicolumn{7}{c}{LLM} \\
\cmidrule(lr){3-10}\cmidrule(lr){11-17}
Method & Metric & V-7 & V-15 & V-23 & V-31 & V-Proj & V-Shallow & V-Deep & V-All & L-06 & L-13 & L-20 & L-27 & L-Shallow & L-Deep & L-All \\
\midrule
One-Class-Recon & Overall & 70.49 & 73.73 & 76.34 & \second{78.67} & 77.84 & 72.66 & 78.33 & \best{79.92} & 74.57 & 71.70 & 72.77 & 71.88 & 73.43 & 72.47 & 73.12 \\
\midrule
One-Class PatchCore & Overall & 67.58 & 69.09 & \best{71.40} & 67.42 & 62.37 & \second{70.14} & 64.01 & 67.42 & 63.52 & 68.41 & 66.99 & 65.36 & 65.73 & 67.82 & 68.94 \\
\bottomrule
\end{tabular}
}
\end{table}

\begin{table}[!htbp]
\centering
\caption{Best feature setting per category on the image anomaly detection datasets.
For each category of MVTec AD and MVTec LOCO AD, the layer setting that achieved the best AUROC with One-Class Recon and One-Class PatchCore is shown.
V-S, V-D, and V-A denote the shallow, deep, and all-layer groups of the visual encoder; L-S, L-D, and L-A denote those of the LLM; V-P denotes the feature after the projector.}
\label{tab:image_category_best}
\setlength{\tabcolsep}{2.4pt}
\renewcommand{\arraystretch}{1.08}
\resizebox{\textwidth}{!}{%
\begin{tabular}{ll*{20}{c}}
\toprule
\multicolumn{2}{c}{} & \multicolumn{15}{c}{MVTec AD} & \multicolumn{5}{c}{MVTec LOCO} \\
\cmidrule(lr){3-17}\cmidrule(lr){18-22}
Method & Metric
& Bottle & Cable & Capsule & Carpet & Grid & Hazelnut & Leather & MetalNut & Pill & Screw & Tile & TBrush & Trans. & Wood & Zipper
& Brkfst & Juice & Screw & Push & Splice \\
\midrule
One-Class-Recon & Layer
& V-S & V-23 & V-S & L-20 & L-D & V-23 & L-D & V-23 & V-23 & V-23 & V-31 & V-S & V-15 & L-A & V-P
& V-D & V-A & V-A & V-S & V-23 \\
& AUROC
& 98.7 & 94.7 & 92.2 & 93.0 & 80.8 & 99.6 & 97.6 & 95.6 & 92.4 & 90.6 & 98.3 & 94.7 & 83.5 & 100 & 86.6
& 88.0 & 93.0 & 76.6 & 79.0 & 82.9 \\
\midrule
One-Class PatchCore & Layer
& V-15 & V-15 & V-15 & V-23 & V-23 & V-23 & V-15 & V-15 & V-23 & V-S & V-23 & V-31 & V-15 & L-27 & V-23
& L-20 & V-A & V-S & V-7 & V-23 \\
& AUROC
& 99.7 & 93.0 & 99.8 & 99.8 & 100 & 99.8 & 100 & 100 & 98.3 & 97.9 & 99.3 & 98.9 & 98.0 & 99.8 & 94.2
& 81.6 & 88.7 & 70.6 & 62.5 & 80.2 \\
\bottomrule
\end{tabular}
}
\end{table}

\subsubsection{Effect of Benchmark Properties}

\paragraph{Video anomaly detection.}
Table~\ref{tab:ubnormal_layer_sweep} shows the results on UBnormal.
V-All was best in the supervised setting and V-Deep in the weakly supervised setting, whereas in the one-class setting L-27 was best with Pred and L-20 with Recon, indicating that deep LLM-side representations were effective.
Since UBnormal contains 29 scenes with diverse actions and anomaly types, representations that abstract the normal context were presumably more stable than low-level visual differences when only normal data are available.
This result differs from the finding of Chen et al.~\cite{chen2025look_shallower} that shallow and intermediate features are effective for vision-centric tasks, showing that deep LLM-side representations can be effective for anomaly detection.

Table~\ref{tab:ucsdped2_layer_sweep} shows the results on UCSD Ped2.
V-All was best in the weakly supervised and one-class Recon settings, indicating that, in contrast to UBnormal, multi-layer visual representations were effective.
UCSD Ped2 consists of a single scene, and its anomalies, such as bicycles and intruding vehicles, are visually distinct; the variation of the normal distribution is therefore small, and compactness was presumably preserved even with multi-layer representations.

\paragraph{Image anomaly detection.}
Table~\ref{tab:mvtecad_layer_sweep_summary} shows the results on MVTec AD.
With Recon, V-Proj outperformed V-All by more than 4 points, and with PatchCore, V-15 was best.
Since MVTec AD mainly contains appearance and structural anomalies, single-layer visual features that suppress low-level variation while retaining part structure were presumably well suited to representing the normal distribution.
On the other hand, in the per-category evaluation in Table~\ref{tab:image_category_best}, the best layer varied widely from the visual side to the LLM side.
This suggests the difficulty of layer selection in anomaly detection and, at the same time, the potential of per-category layer selection.

Table~\ref{tab:mvtecloco_layer_sweep_summary} shows the results on MVTec LOCO AD.
With Recon, V-All was best, indicating that, unlike MVTec AD, multi-layer visual representations were effective.
Since LOCO contains logical anomalies, i.e., inconsistencies in counts and combinations, this suggests the effectiveness of multi-layer, deep visual representations that can capture both local structure and global configuration.
Moreover, the fact that LLM features were rarely best on MVTec suggests that linguistic abstraction may weaken local appearance anomalies.

\subsubsection{Effect of Supervision and Classifiers}

\paragraph{Video anomaly detection.}
In the supervised setting, V-All was best on UBnormal and L-13 on UCSD Ped2, so the optimal layer was not uniform.
Nevertheless, the effectiveness of multi-layer visual representations on UBnormal suggests that supervised classification can exploit features at multiple granularities, from appearance differences to semantic context.

In the weakly supervised setting, V-Deep was best on UBnormal and V-All on UCSD Ped2, indicating that deep or multi-layer visual features were effective.
In particular, on UBnormal, V-Deep outperformed V-Shallow by more than 13 points.
A likely reason is that video-level labels alone cannot directly supervise local appearance, so deep visual features that encode scenes, actions, and object relations align more easily with the labels.

In the one-class setting, an LLM-side single layer was best on UBnormal and visual-side features were best on UCSD Ped2, suggesting the importance of a compact normal manifold.
We also attribute the higher performance of Recon over Pred to the fact that feeding video segments to the LVLM averages out local changes within frames; reconstruction is therefore better suited than predicting the next segment from motion states.

\paragraph{Image anomaly detection.}
With Recon, V-Proj was best on MVTec AD and V-All on MVTec LOCO AD, so the optimal representation depended on whether anomalies are appearance-based or logical.
That is, a compact single-layer representation was effective on MVTec AD, whereas a multi-layer visual representation that handles global--local consistency was effective on LOCO.

With PatchCore, in the per-category evaluation in Table~\ref{tab:image_category_best}, an intermediate single visual layer was best for almost all categories of MVTec AD, while multi-layer features were best for some categories of LOCO.
Multi-layer features can thus be effective for logical anomalies that require global consistency.
On the other hand, PatchCore on LOCO was about 8 points lower than Recon, suggesting that local nearest-neighbor distances have difficulty capturing inconsistencies in the overall layout, counts, and combinations of an image.

\section{Conclusion}

We analyzed effective representations for image and video anomaly detection using the multi-layer visual representations of large vision-language models.
Our experiments showed that, in contrast to prior claims that shallow/intermediate layers and multi-layer fusion are effective~\cite{chen2025look_shallower,lin2025multi_layer_visual_feature_fusion}, the effective representation for anomaly detection depends on the properties of the normal distribution and on the supervision, and can be a single layer or a deep LLM-side feature.
As future work, we will investigate new anomaly detection methods that adaptively select and fuse layer representations according to the anomaly type and input characteristics.

\bibliographystyle{plain}
\bibliography{refs}

\end{document}